\documentclass{article}
\usepackage{spconf,amsmath,graphicx,hyperref}
\usepackage{booktabs}
\usepackage{tabularx}
\usepackage{makecell}
\usepackage{multirow}
\usepackage{subcaption}
\title{Triage: Hierarchical Visual Budgeting for Efficient Video Reasoning in Vision-Language Models}
\name{
    Anmin Wang$^{1}$, 
    Nan Zhang$^{2*}$, 
    Wei Tao$^{1, 2}$,
    Xiaoyang Qu$^{2}$,
    Guokuan Li$^{1*}$,
    Jiguang Wan$^{1}$,
    Jianzong Wang$^{2}$
    \thanks{*Corresponding authors.}
}
\address{
    $^{1}$Huazhong University of Science and Technology, Wuhan, China \\
    $^{2}$Ping An Technology (Shenzhen) Co., Ltd., Shenzhen, China \\
}

\begin{document}
%\ninept
%
\maketitle
\begin{abstract}
Vision-Language Models (VLMs) face significant computational challenges in video processing due to massive data redundancy, which creates prohibitively long token sequences. To address this, we introduce Triage, a training-free, plug-and-play framework that reframes video reasoning as a resource allocation problem via hierarchical visual budgeting. Its first stage, Frame-Level Budgeting, identifies keyframes by evaluating their visual dynamics and relevance, generating a strategic prior based on their importance scores. Guided by this prior, the second stage, Token-Level Budgeting, allocates tokens in two phases: it first secures high-relevance Core Tokens, followed by diverse Context Tokens selected with an efficient batched Maximal Marginal Relevance (MMR) algorithm. Extensive experiments demonstrate that Triage improves inference speed and reduces memory footprint, while maintaining or surpassing the performance of baselines and other methods on various video reasoning benchmarks.
\end{abstract}
\begin{keywords}
Vision-Language Models, Video Reasoning, Efficient Inference, Token Selection
\end{keywords}
\section{Introduction}
\label{sec:intro}

Vision-Language Models (VLMs) \cite{radford2021learning, liu2024improved} have revolutionized how machines interpret the visual world, enabling complex reasoning and conversations about visual content. Although many VLMs can now process video \cite{li2025videochat, cheng2024videollama2advancingspatialtemporal}, the ability to efficiently and effectively reason over video remains a challenge \cite{ding2025dolanguage}. Unlike images, videos introduce the temporal dimension, encoding event structures, causal relationships, and state transitions, making video comprehension vital for next-generation applications. However, current architectures suffer from high computational and memory costs \cite{shinde2025surveyefficientvisionlanguagemodels, tu2025vlcache}. This stems from two main problems. The first is temporal and information redundancy, where many frames are either highly similar to their neighbors or irrelevant to the user's query. The second is spatial redundancy, where tokens within a relevant frame are either redundant background or highly similar to adjacent ones. These redundancies inflate the Key-Value (KV) cache, leading to excessive memory usage. 
% Traditional VLMs encounter OOM (Out of Memory) problem beyond 128 frames under a 24GB GPU memory limit and the oversized KV cache also increases inference latency.

% \begin{figure*}[t]
% \centering
% \includegraphics[width=0.95\linewidth]{pictures/transfer-s.pdf} % Reduce the figure size so that it is slightly narrower than the column.
% \caption{Motivation for Triage. (a) Video has temporal and spatial redundancy, which leads to (b) a memory bottleneck. (c) Our solution uses hierarchical budgeting to select key information for efficient reasoning.}
% \label{fig:motivation}
% \end{figure*}

Several strategies have been proposed to mitigate these issues. Frame-level methods have evolved from uniform sampling to adaptive methods that identify and discard irrelevant frames \cite{liu2025boltboostlargevisionlanguage}. Token-level methods include token pruning, which removes less important visual tokens \cite{tao2025dycokedynamiccompressiontokens, xing2025pyramiddropacceleratinglargevisionlanguage, chen2024fastv, liu2025videocompressioncommanderplugandplay}, and KV cache compression \cite{tu2025vlcache, wang2025adaretakeadaptiveredundancyreduction}, which reduces memory footprint during decoding. However, these methods often operate disjointly. Few works approach this as a unified, hierarchical budgeting problem where coarse-grained frame selection can proactively guide fine-grained token allocation.

To overcome these limitations, we propose Triage, a training-free, plug-and-play framework that addresses redundancy through a hierarchical visual budgeting approach. Inspired by medical triage, it treats inference as a resource allocation problem by implementing a hierarchical visual budgeting process. Triage implements a synergistic two-stage process to allocate resources to the most critical visual information, significantly reducing the token count for more efficient and robust video reasoning.

The first stage, Frame-Level Budgeting, executes a coarse-grained budget to tackle temporal and information redundancy by computing a frame importance score for candidate frames based on visual dynamics and relevance. It then uses an Adaptive Temporal Bucketing strategy to select keyframes, balancing informational value with broad temporal coverage. Crucially, the scores of keyframes are preserved, forming the prior that guides the subsequent token-level allocation.

% \begin{figure*}[t]
% \centering
% \includegraphics[width=0.98\linewidth]{pictures/main_pic_2.pdf} % Reduce the figure size so that it is slightly narrower than the column.
% \caption{An overview of the Triage framework. (a) Frame-Level Budgeting selects important keyframes, generating a strategic prior. (b) Token-Level Budgeting then uses this prior to select a compact and informative token sequence.}
% \label{fig:method}
% \end{figure*}

\begin{figure*}[t]
\centering
\includegraphics[width=0.98\linewidth]{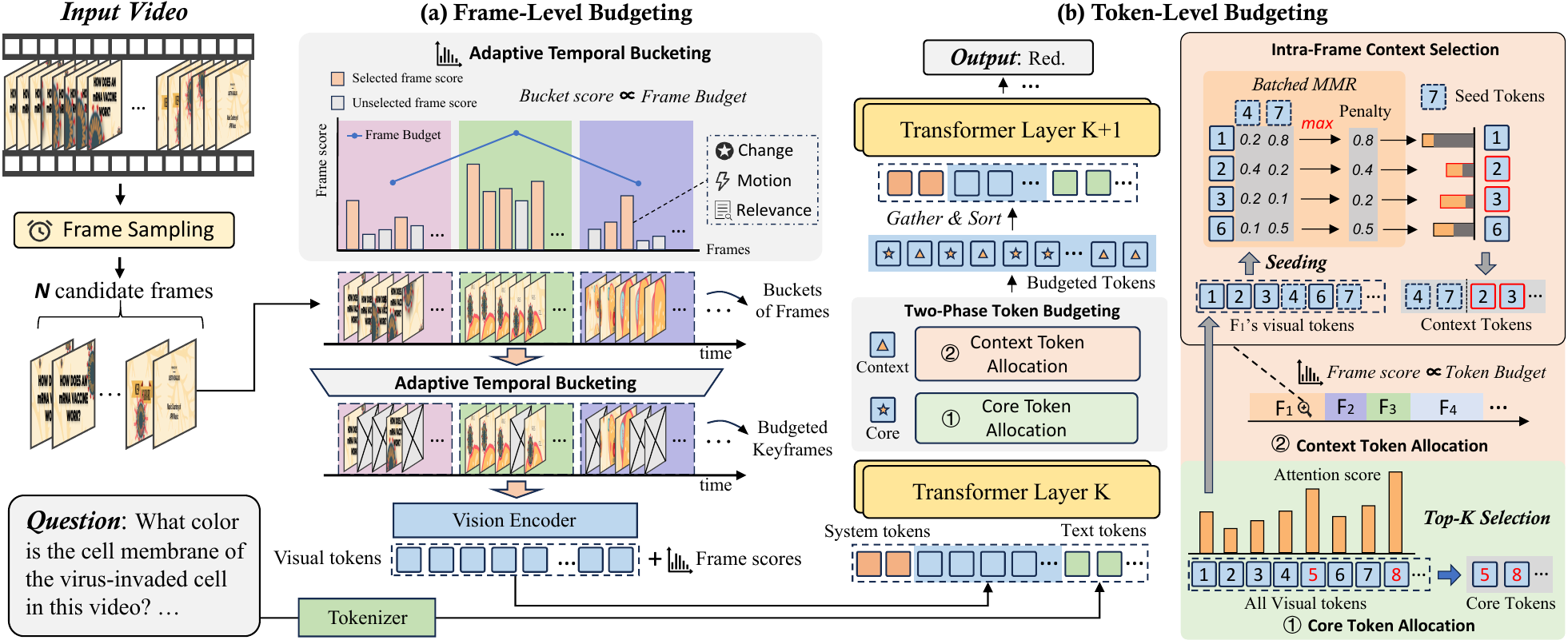} % Reduce the figure size so that it is slightly narrower than the column.
\caption{An overview of the Triage framework.}
\label{fig:method}
\end{figure*}

The second stage, Token-Level Budgeting, then executes a fine-grained, two-phase budget to address spatial redundancy within each selected keyframe. Guided by the strategic prior from the first stage, this process determines how to spend its allocated token budget to capture the most valuable information. It employs a two-phase allocation scheme to maximize informational value: (1) Core Token Allocation secures a small set of tokens with the highest query relevance. (2) Context Token Allocation then uses the remaining budget to select a diverse set of tokens that provide broader context, using a batched Maximal Marginal Relevance (MMR) algorithm \cite{carbonell1998use}. Extensive experiments demonstrate that our Triage framework significantly improves inference efficiency while maintaining or even improving performance on video reasoning tasks.

\section{Proposed method}
% \subsection{Overall Framework}
% \label{sec:overall_framework}
% As illustrated in Figure \ref{fig:method}, our Triage framework reframes video reasoning as a hierarchical resource allocation problem, operating in two stages. First, Frame-Level Budgeting performs a coarse-grained analysis to select a compact set of keyframes based on their importance scores, yielding a guiding prior. The second stage, Token-Level Budgeting, then executes a fine-grained token selection within these keyframes. Guided by this prior, the second stage, Token-Level Budgeting, executes a fine-grained selection within these keyframes. It proportionally distributes the token budget based on frame scores and then employs an MMR-based algorithm to build a final, diverse token set, ensuring the computational budget is spent on maximally informative content.

\subsection{Frame-Level Budgeting}
\label{sec:frame_selection}
As the first stage of our framework, Frame-Level Budgeting tackles temporal and information redundancy by selecting a compact set of keyframes without invoking the VLM's vision encoder. It begins by uniformly sampling an initial candidate set of $N$ frames, $\{f_1, f_2, \dots, f_N\}$, with the objective of selecting a smaller final set of $M$ keyframes ($M \ll N$).

% \subsubsection{Unified Frame Scoring Mechanism}
\textbf{Frame Importance Scoring.} To select keyframes, we calculate a frame importance score to evaluate the importance of each frame. The resulting frame importance score, $S_{\text{frame}}$, is designed to holistically balance a frame's visual dynamics (scene change and motion intensity) with its semantic relevance to the query.

For a given frame $f_i$ and query $Q$, the frame score is calculated as a weighted sum of three distinct components:
\begin{equation}
\label{eq:frame_score}
\begin{split}
S_{\text{frame}}(f_i, Q) = w_c S_{\text{change}}(f_i) &+ w_m S_{\text{motion}}(f_i) \\
&+ w_r S_{\text{relevance}}(f_i, Q),
\end{split}
\end{equation}
where $w_c, w_m,$ and $w_r$ are balancing hyperparameters. Each component captures an aspect of a frame's value:

\textit{Scene Change.} This component is calculated as the inverse cosine similarity between adjacent frames, identifying temporal shifts like scene cuts to filter static content.
% \begin{equation}
%     \label{eq:change}
%     S_{\text{change}}(f_i) = 1 - \frac{\mathbf{p}_i \cdot \mathbf{p}_{i-1}}{\|\mathbf{p}_i\|\| \mathbf{p}_{i-1}\|}.
% \end{equation}

\textit{Motion Intensity.} Calculated as the L2 norm of the pixel-wise difference between frames, this component captures the magnitude of movement to distinguish high-action segments.
% \begin{equation}
%     \label{eq:motion}
%     S_{\text{motion}}(f_i) = \|\mathbf{p}_i - \mathbf{p}_{i-1}\|_2.
% \end{equation}

\textit{Text Relevance.} This component is the cosine similarity between the frame's and query's embeddings, computed by a lightweight model like CLIP to align with the user's query.
% \begin{equation}
%     \label{eq:relevance}
%     S_{\text{relevance}}(f_i, Q) = \frac{E_{f_i} \cdot E_{Q}}{\|E_{f_i}\|\|E_{Q}\|}.
% \end{equation}

\textbf{Adaptive Temporal Bucketing.} To avoid clustering keyframes and preserve the video's narrative, we employ Adaptive Temporal Bucketing strategy. First, we divide the $N$ candidate frames into $K$ chronological buckets and calculate a "bucket score" $W_k$ for each bucket $B_k$ by summing its internal frame scores, where $W_k = \sum_{f_i \in B_k} S_{\text{frame}}(f_i)$. 
% A higher bucket score indicates that a bucket contains a greater concentration of visually dynamic or query-relevant moments, thus justifying a larger portion of the frame budget. 

To allocate the budget of $M$ keyframes, a baseline allocation of one frame is first assigned to each of the $K$ buckets to ensure full temporal coverage. Then the remaining budget of $M-K$ frames are allocated among the buckets in proportion to their bucket scores $W_k$. Finally, from each bucket $B_k$, we select the top-$n_k$ frames with the highest $S_{\text{frame}}$ scores, where $n_k$ is the total number of frames allocated to the bucket. This process yields the budgeted keyframes and their frame importance scores, which form the strategic prior for the next stage.

\subsection{Token-Level Budgeting}
\label{sec:token_budgeting}
% Following the coarse-grained frame budgeting, this second stage executes a fine-grained budget on visual tokens from the selected $M$ keyframes, aiming to construct an highly informative token sequence within a total budget $B_T$. The allocation is guided by both the frame importance scores and token importance scores derived from query interaction. It unfolds in two distinct and complementary phases: an initial allocation for Core Tokens with the highest query relevance, followed by a more sophisticated, prior-guided allocation for a diverse set of Context Tokens.

Following the coarse-grained frame budgeting, this second stage performs fine-grained budgeting on visual tokens from the selected $M$ keyframes, aiming to construct a highly informative token sequence within a total budget $B_T$. % The allocation is guided by both the frame importance scores and token importance scores derived from query interaction. This process has two complementary phases: an initial allocation for highly relevant Core Tokens, followed by a prior-guided allocation for a diverse set of Context Tokens.

\textbf{Token Importance Scoring.} To guide the token budgeting process, we quantify the importance of each visual token in relation to the text query. This is achieved by leveraging the cross-attention scores from a designated Transformer layer with $H$ heads.

We obtain a cross-attention matrix between the $N_q$ query tokens and the $N_v$ visual tokens and the importance score of a visual token is then computed by averaging its attention weights across all query tokens and attention heads, i.e., $S_{\text{token}}(j) = \tfrac{1}{N_q H}\sum_{i,h} A_{ij}^{(h)}$, where $A_{ij}^{(h)}$ is the attention from query token $i$ to visual token $j$ under head $h$. This score provides a context-dependent measure that serves as the basis for token allocation.

% With the token importance score computed for each token, the total token budget $B_T$ is then spent in two phases to ensure both relevance and diversity. The first phase acts as a conservative safeguard for the most query-relevant information, while the second phase uses the remaining budget more exploratively to select a broader, more diverse set of informative tokens.
\textbf{Two-Phase Token Budgeting.} With the token importance score computed, the total token budget $B_T$ is spent in two phases to ensure both relevance and diversity. The first phase secures the most query-relevant information as a safeguard, while the second phase uses the remaining budget to select a diverse set of informative tokens.

\textit{Phase 1: Core Token Allocation.}
% The first phase involves a non-negotiable expenditure to secure a small fixed-ratio set of Core Tokens. This allocation is guided purely by semantic relevance to the query. We allocate a budget of $B_{\text{core}}$ tokens and spend it on those with the highest token importance scores, preserving the most critical visual evidence. This core set $\mathcal{T}_{\text{core}}$ is formed by:
% \begin{equation}
% \label{eq:core_selection}
% \mathcal{T}_{\text{core}} = \underset{j \in \{1, \dots, N_v\}}{\operatorname{arg~top}_{B_{\text{core}}}} (S_{\text{token}}(j)).
% \end{equation}
% This step acts as a safeguard, guaranteeing a baseline of high-relevance information is passed to the model. This preemptive selection is critical because it ensures that the most direct visual answers to the query are protected from being discarded by the subsequent diversity-aware stage.
The first phase reserves a fixed-ratio budget of $B_{\text{core}}$ to form Core Tokens, selected purely by their relevance to the query. This allocation prioritizes tokens with the highest importance scores, preserving critical visual evidence. The set is defined as follows:

\begin{equation}
\label{eq:core_selection}
\mathcal{T}_{\text{core}} = \underset{j \in \{1, \dots, N_v\}}{\operatorname{arg~top}_{B_{\text{core}}}} (S_{\text{token}}(j)).
\end{equation}

This step acts as a safeguard, guaranteeing that the most direct visual answers to the query are preserved before the subsequent diversity-aware selection stage.

% \textbf{Phase 2: Context Token Allocation.}
% After securing the Core Tokens, the remaining budget $B_{\text{context}} = B_T - B_{\text{core}}$ is spent on selecting a diverse set of Context Tokens. While Core Tokens anchor the model's focus on primary subjects, these tokens are crucial for capturing the broader visual context (e.g., surrounding environments, object interactions) needed for robust reasoning. To allocate this budget effectively, we employ a two-step process. First, Inter-Frame Budget Distribution divides the total budget $B_{\text{context}}$ among the selected keyframes based on their frame importance scores. Second, Intra-Frame Context Selection uses each frame's assigned budget to select a final, diverse set of tokens to mitigate spatial redundancy within the frame.

% First, in Inter-Frame Budget Distribution, the token budget $B_{\text{context}}$ is distributed among the $M$ keyframes, guided by the frame importance scores from the first stage. This prior consists of the normalized final importance scores ($S_{\text{frame}}$) calculated for each selected keyframe. The budget $B(f)$ allocated to each frame $f$ is proportional to this score:
% \begin{equation}
% \label{eq:token_budget_allocation}
% B(f) = \text{round} \left( B_{\text{context}} \times \frac{S_{\text{frame}}(f)}{\sum_{k=1}^{M} S_{\text{frame}}(k)} \right).
% \end{equation}

\textit{Phase 2: Context Token Allocation.}
After securing Core Tokens, the remaining budget $B_{\text{context}} = B_T - B_{\text{core}}$ is spent on selecting diverse Context Tokens to capture broader visual context (e.g., object interactions) needed for robust reasoning. This allocation is a two-step process. First, Inter-Frame Budget Distribution divides the budget $B_{\text{context}}$ among keyframes based on their frame importance scores. Second, Intra-Frame Context Selection uses each frame's assigned budget to select a final, diverse set of tokens to mitigate spatial redundancy.

First, in Inter-Frame Budget Distribution, the token budget $B_{\text{context}}$ is distributed among the $M$ keyframes. The budget $B(f)$ allocated to each frame $f$ is directly proportional to its frame importance score ($S_{\text{frame}}(f)$).
% \begin{equation}
% \label{eq:token_budget_allocation}
% B(f) = \text{round} \left( B_{\text{context}} \times \frac{S_{\text{frame}}(f)}{\sum_{k=1}^{M} S_{\text{frame}}(k)} \right).
% \end{equation}

% Second, in Intra-Frame Context Selection, we spend the budget $B(f)$ to select a diverse set of Context Tokens within each frame. A naive approach of simply picking the tokens with the next highest $S_{\text{token}}$ scores would likely result in a visually homogeneous set. To overcome this, we incorporate a diversity-aware selection mechanism by adapting the Maximal Marginal Relevance (MMR) principle, an algorithm originally pioneered in the fields of text summarization and information retrieval. However, a classic, iterative MMR implementation is computationally prohibitive for real-time inference. Therefore we introduce an efficient, batched approximation of the MMR objective. The batched process begins by selecting a small number ($k_s$) of high-relevance seed tokens ($\mathcal{T}_{\text{seed}}$) from the set of non-core candidates in frame $f$, denoted $\mathcal{C}_f$. Following this, an additional set of $m = B(f) - |\mathcal{T}_{\text{seed}}|$ tokens, denoted $\mathcal{T}_{\text{context}}$, is selected from the remaining candidates based on the following diversity-aware selection criterion:

\begin{table*}[t]
    \centering
    \small
    \newcolumntype{C}[1]{>{\centering\arraybackslash}p{#1}}
    \renewcommand{\arraystretch}{0.57}
    % \begin{tabularx}{\textwidth}{ l l >{\Centering}p{1.6cm} C{0.65cm} C{0.65cm} C{0.65cm} C{0.8cm} *{3}{>{\Centering}X} }
    \caption{Comprehensive performance evaluation on four video reasoning benchmarks.}
    \begin{tabular*}{\textwidth}{@{\extracolsep{\fill}} l l ccccc ccc}
    \toprule
    
    % --- 表头结构 ---
    \multirow{3}{*}{\textbf{Model}} 
    & \multirow{3}{*}{\textbf{Method}} 
    & \multirow{3}{*}{\makecell[c]{\textbf{Retention}\\\textbf{Ratio}}} 
    & \multicolumn{4}{c}{\textbf{Video-MME}} 
    & \multirow{3}{*}{\makecell[c]{\textbf{LongVideo}\\\textbf{Bench}}} 
    & \multirow{3}{*}{\textbf{MVBench}} 
    & \multirow{3}{*}{\textbf{LVBench}} \\
    \cmidrule(lr){4-7}
    & & 
    & \small\text{short} & \small\text{medium} & \small\text{long} & \small\text{overall} 
    & & & \\
    \midrule
    
    % --- LLaVA-OneVision 数据 ---
    \multirow{10}{*}{\makecell[c]{LLaVA-\\OneVision-7B}} 
    & Vanilla & 100\% & 71.2 & 54.2 & 48.1 & 58.7 & 56.7 & 55.8 & 40.3 \\
    \cmidrule(l){2-10}
    & PyramidDrop & 50\% & 71.0 & 57.5 & 47.3 & 58.6 & 57.1 & 55.7 & 40.8 \\
    & FastV & 50\% & 71.8 & 57.6 & 47.8 & 59.0 & 56.4 & 55.7 & 40.4 \\
    & DyCoke & 50\% & 71.6 & 56.0 & 49.7 & 59.1 & 56.3 & 55.3 & 40.4 \\
    \cmidrule(l){2-10} 
    & \multirow{4}{*}{Triage (Ours)} 
    & 100\% & 72.3 & 55.0 & 50.8 & 59.4 & 58.3 & 55.9 & 41.7 \\
    & & 75\% & 71.5 & 56.5 & \textbf{51.0} & 59.7 & 57.9 & \textbf{56.1} & 42.0 \\
    & & 50\% & \textbf{72.4} & \textbf{56.8} & 50.8 & \textbf{60.1} & 57.6 & 55.7 & \textbf{43.3} \\
    & & 25\% & \textbf{72.4} & 56.8 & 50.7 & 60.0 & \textbf{58.6} & 55.5 & 41.1 \\
    \midrule
    
    % --- LLaVA-Video 数据 ---
    \multirow{10}{*}{\makecell[c]{LLaVA-\\Video-7B}}
    & Vanilla & 100\% & 71.0 & 55.1 & 48.9 & 58.3 & 56.8 & 58.9 & 36.8 \\
    \cmidrule(l){2-10}
    & PyramidDrop & 50\% & 70.0 & 54.1 & 51.1 & 58.4 & 56.2 & 58.2 & 36.5 \\
    & FastV & 50\% & 70.2 & 54.7 & \textbf{51.5} & 58.7 & 54.0 & 58.0 & 36.2 \\
    & DyCoke & 50\% & 70.8 & 54.1 & 49.8 & 58.1 & 56.1 & 58.0 & 35.4 \\
    \cmidrule(l){2-10}
    & \multirow{4}{*}{Triage (Ours)} & 100\% & \textbf{71.7} & \textbf{56.0} & 49.6 & \textbf{59.1} & \textbf{59.0} & \textbf{59.2} & 39.9 \\
    & & 75\% & 70.8 & 54.7 & 48.8 & 58.1 & 58.3 & 58.1 & 40.4 \\
    & & 50\% & 71.0 & 55.1 & 50.2 & 58.8 & 58.0 & 58.3 & \textbf{40.5} \\
    & & 25\% & 68.0 & 52.2 & 48.1 & 56.1 & 55.0 & 55.1 & 39.3 \\
    \midrule
    
    % --- Qwen2-VL 数据 ---
    \multirow{10}{*}{\makecell[c]{Qwen2-VL-7B}}
    & Vanilla & 100\% & 70.8 & \textbf{58.2} & 49.7 & 59.6 & 55.7 & 64.5 & 38.7 \\
    \cmidrule(l){2-10}
    & PyramidDrop & 50\% & 67.6 & 54.2 & 49.6 & 57.1 & 54.6 & 60.0 & 38.9 \\
    & FastV & 50\% & 66.9 & 53.7 & 49.0 & 56.6 & 52.8 & 58.5 & 37.3 \\
    & DyCoke & 50\% & 67.9 & 53.7 & 49.0 & 56.9 & 52.7 & 58.5 & 37.7 \\
    \cmidrule(l){2-10}
    & \multirow{4}{*}{Triage (Ours)} & 100\% & \textbf{72.1} & 57.9 & \textbf{52.1} & \textbf{60.7} & 56.7 & \textbf{64.7} & 40.1 \\
    & & 75\% & 71.3 & 57.2 & 50.3 & 59.7 & \textbf{57.0} & 63.8 & 40.8 \\
    & & 50\% & 70.7 & 57.2 & 49.9 & 59.3 & 55.9 & 63.0 & \textbf{41.1} \\
    & & 25\% & 66.3 & 55.4 & 48.7 & 56.8 & 54.8 & 63.7 & 38.9 \\
    \bottomrule 
    \end{tabular*}
    \label{tab:comparison_results}
\end{table*}

Second, in Intra-Frame Context Selection, we spend the budget $B(f)$ to select  diverse Context Tokens within each frame. To avoid selecting a visually homogeneous set, we adapt the Maximal Marginal Relevance (MMR) principle, an algorithm originally pioneered in text summarization and information retrieval. As classic MMR implementation is computationally prohibitive for real-time inference,  we introduce an efficient, batched approximation. The batched process first selects a small number ($k_s$) of high-relevance seed tokens ($\mathcal{T}_{\text{seed}}$) from non-core token candidates $\mathcal{C}_f$ in frame $f$. Then, the remaining $m = B(f) - |\mathcal{T}_{\text{seed}}|$ tokens form $\mathcal{T}_{\text{context}}$ by:
\begin{equation}
\label{eq:context_selection}
\mathcal{T}_{\text{context}} = \underset{t_i \in \mathcal{C}_f \setminus \mathcal{T}_{\text{seed}}}{\operatorname{arg~top}_{m}} \left(S_{\text{token}}(t_i) - \lambda \max_{t_s \in \mathcal{T}_{\text{seed}}} \text{sim}(v_i, v_s) \right),
\end{equation}
where $\lambda$ balances relevance and diversity, $\text{sim}(v_i, v_s)$ is the cosine similarity between key states. The complete set of Context Tokens for frame $f$ is the union of $\mathcal{T}_{\text{seed}}$ and $\mathcal{T}_{\text{context}}$.

Ultimately, the final visual token set $\mathcal{T}_{\text{final}} = \mathcal{T}_{\text{core}} \cup (\bigcup_f \mathcal{T}_{\text{seed}}(f)) \cup (\bigcup_f \mathcal{T}_{\text{context}}(f))$ forms a compact yet highly informative sequence. This sequence is then fed into the subsequent layers, achieving significant computational savings.

% \begin{figure*}[htb]
%   \centering
%   \small
  
%   \begin{subfigure}[t]{0.33\textwidth} % 使用 subfigure 环境，[t] 仍然表示顶部对齐
%     \centering
%     \includegraphics[width=\linewidth]{pictures/inference_time_comparison.pdf}
%     \caption{Result 1} % 使用 \caption 生成子图标题
%     \label{fig:res1} % 可以为子图添加单独的标签
%   \end{subfigure}
%   \hfill % \hfill 仍然用来分配间距
%   \begin{subfigure}[t]{0.33\textwidth}
%     \centering
%     \includegraphics[width=\linewidth]{pictures/inference_time_comparison.pdf}
%     \caption{Result 3}
%     \label{fig:res3}
%   \end{subfigure}
%   \hfill
%   \begin{subfigure}[t]{0.29\textwidth}
%     \centering
%     \includegraphics[width=\linewidth]{pictures/hyperparameter_sensitivity.pdf}
%     \caption{Result 4}
%     \label{fig:res4}
%   \end{subfigure}

%   \caption{Example of placing three figures side by side across two columns.} % 这是整个图的总标题
%   \label{fig:res}
% \end{figure*}

\section{Experiments}

\subsection{Experimental Setup}

% To comprehensively evaluate our proposed method, we conduct experiments using various VLMs and a diverse set of video reasoning benchmarks, covering a wide range of video understanding tasks and durations.

\textbf{Models and Benchmarks.} To demonstrate the plug-and-play nature of our framework, we integrate it into three distinct open-source VLMs: LLaVA-OneVision-7B \cite{li2024llavaonevisioneasyvisualtask}, LLaVA-Video-7B \cite{zhang2024llavavideoinstructiontuningsynthetic} and Qwen2-VL-7B \cite{wang2024qwen2vlenhancingvisionlanguagemodels}. 
Our evaluation is performed on four video reasoning benchmarks: Video-MME \cite{fu2025videomme}, MVBench \cite{li2024mvbench}, LongVideoBench \cite{wu2024longvideobench}, LVBench \cite{wang2024lvbenchextremelongvideo}.

\textbf{Baselines.} We compare Triage with two types of baselines: (1) the vanilla baseline, which is the original VLM that processes all visual tokens, serving as the reference for performance and efficiency; and (2) other token compression methods: PyramidDrop \cite{xing2025pyramiddropacceleratinglargevisionlanguage}, FastV \cite{chen2024fastv} and DyCoke \cite{tao2025dycokedynamiccompressiontokens}. The results of all baseline methods were obtained by reproducing their methodologies within a unified experimental setup. 

% \textbf{Configurations.} Our training-free method is evaluated at multiple token retention ratios (which set the total token budget $B_T$), including 100\%, 75\%, 50\%, and 25\%. The results of all baseline methods were obtained by reproducing their methodologies within a unified experimental setup. 

% \begin{figure}[t]
% \centering
% \includegraphics[width=0.98\columnwidth]{pictures/performance_efficiency_tradeoff.pdf} % Reduce the figure size so that it is slightly narrower than the column. Don't use precise values for figure width.This setup will avoid overfull boxes.
% \caption{Analysis between performance and efficiency.}
% \label{fig:tradeoff}
% \end{figure}

\subsection{Main Results}

\textbf{Comparison with Vanilla Models.} As shown in Table \ref{tab:comparison_results}, Triage consistently outperforms the vanilla baselines. With only the Frame-Level Budgeting applied (100\% retention ratio), performance improves across all models. When further applying Token-Level Budgeting, Triage maintains highly competitive performance and sometimes achieves additional gains even at reduced retention ratios. This trend is particularly evident on LVBench, where all models achieve their highest scores at a 50\% retention ratio.

% \subsubsection{Comparison with Other Methods}

\textbf{Comparison with Other Methods.} We also compare Triage with several efficiency-enhancing methods. At a comparable 50\% token retention ratio, Triage generally achieves better performance. For example, on Video-MME with LLaVA-OneVision-7B, Triage achieves an overall score of 60.4, outperforming PyramidDrop (58.6), FastV (59.0), and DyCoke (59.1). The performance margin is particularly clear on LVBench dataset, where Triage consistently outperforms all other methods across all three base models.

Triage's strong performance-efficiency trade-off is  demonstrated at lower retention rates. Notably, Triage at a 25\% budget outperforms others at 50\%. For instance, with LLaVA-OneVision-7B, its score on LongVideoBench at 25\% retention surpasses all competing methods at double the budget.

\begin{table}[t]
\setlength{\abovecaptionskip}{2pt}
\caption{Ablation study results.}
\label{tab:ablation}
\begin{center}
\small
\renewcommand{\arraystretch}{0.8}
\begin{tabular}{lc}
\toprule
\textbf{Configuration} & \textbf{Video-MME} \\
\midrule
\textbf{Triage (Full Method)} & \textbf{60.1} \\
\midrule
% --- Ablation group for the overall framework ---
\multicolumn{2}{l}{\textit{Ablations on Overall Framework:}} \\ 
\quad w/o Frame-Level Budgeting & 58.3 \\
\quad w/o Token-Level Budgeting & 59.4 \\
\midrule
% --- Ablation group for key mechanisms ---
\multicolumn{2}{l}{\textit{Ablations on Key Mechanisms:}} \\
\quad w/o Adaptive Temporal Bucketing & 58.3 \\
% \quad w/o Inter-Frame Budget Distribution & 59.5 \\
\quad w/o Core Token Allocation & 59.4 \\
\quad w/o Context Token Allocation & 59.3 \\
\bottomrule
\end{tabular}
\end{center}
\end{table}

\begin{figure}[htb]

\begin{minipage}[b]{0.39\linewidth}
  \centering
  \small
  \centerline{\includegraphics[width=3.4cm]{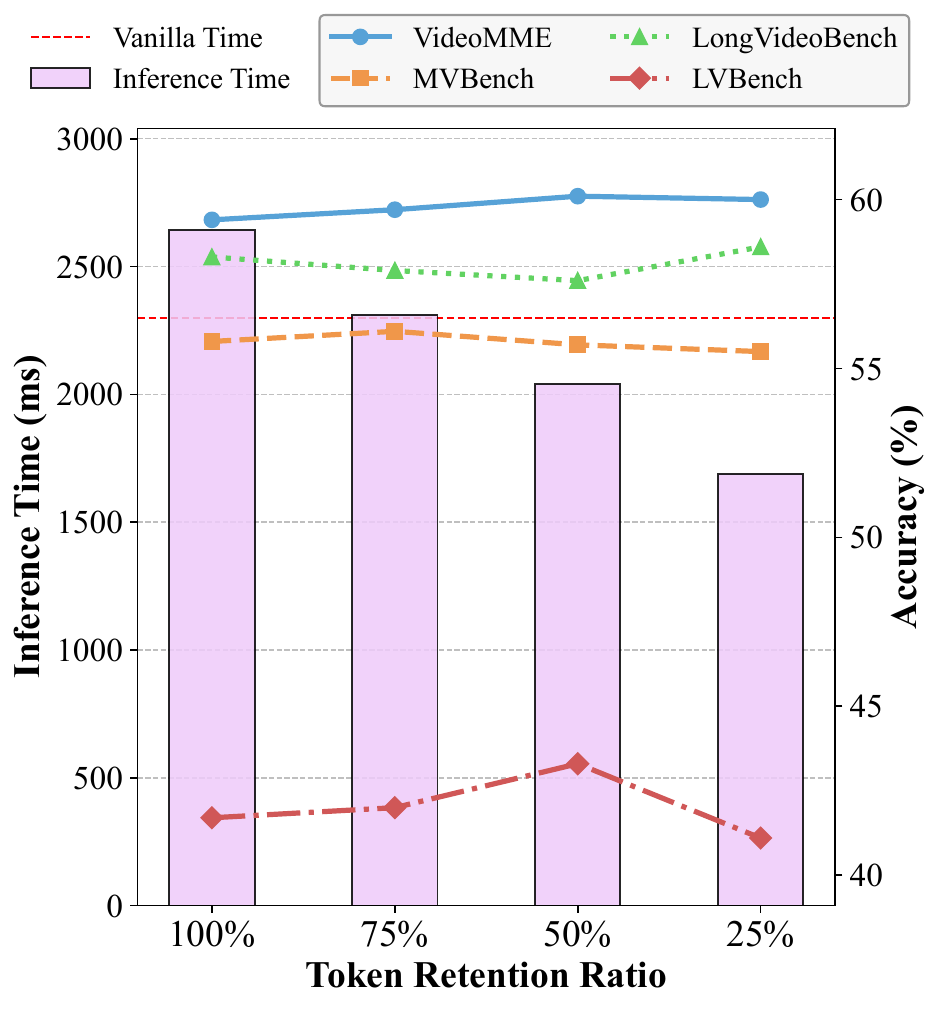}}
%  \vspace{1.5cm}
  \centerline{(a) Trade-off}\medskip
\end{minipage}
\hfill
\begin{minipage}[b]{0.58\linewidth}
  \centering
  \small
  \centerline{\includegraphics[width=4.9cm]{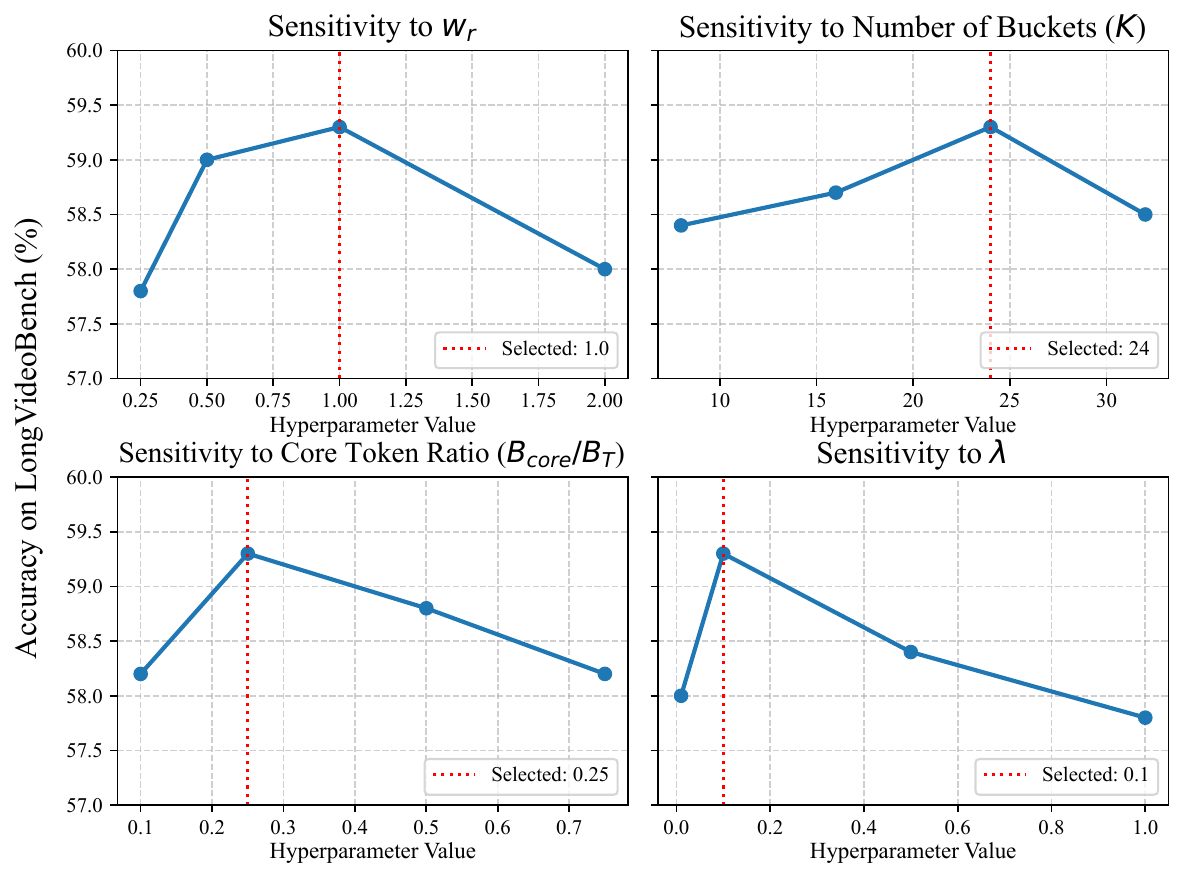}}
%  \vspace{1.5cm}
  \centerline{(b) Hyperparameter Sensitivity}\medskip
\end{minipage}
\caption{Performance-efficiency trade-off and hyperparameter sensitivity analysis for Triage.}
\label{fig:res}
\end{figure}

\subsection{Ablation Study and Analysis}

% \begin{figure}[htb]

% \begin{minipage}[b]{0.48\linewidth}
%   \small
%   \centering
%   \centerline{\includegraphics[width=3.5cm]{image3}}
% %  \vspace{1.5cm}
%   \centerline{(a) Inference Time}\medskip
% \end{minipage}
% \hfill
% \begin{minipage}[b]{0.48\linewidth}
%   \small
%   \centering
%   \centerline{\includegraphics[width=5.2cm]{image4}}
% %  \vspace{1.5cm}
%   \centerline{(b) Hyperparameter Sensitivity}\medskip
% \end{minipage}
% %
% \caption{Experimental results of inference time and sensitivity.}
% \label{fig:res}
% %
% \end{figure}

% \textbf{Ablation Study.} We conduct ablation studies on Video-MME using the LLaVA-OneVision-7B model to quantify each component's contribution (Table \ref{tab:ablation}). Removing either main stage—Frame-Level Budgeting or Token-Level Budgeting—causes a significant performance drop. This demonstrates that Frame-Level Budgeting is critical for eliminating irrelevant frames, and Token-Level Budgeting actively enhances reasoning by strategically removing redundant tokens.
% Similarly, ablating key internal mechanisms also degrades performance. Disabling Inter-Frame Budget Distribution causes a notable performance drop, demonstrating proportional budget allocation is more effective than naive uniform allocation. Furthermore, ablating either Core Token Allocation or Context Token Allocation degrades performance, showing that Core Tokens act as an indispensable anchor for query relevance, while Context Tokens are crucial for providing diverse visual information for robust video reasoning.

\textbf{Ablation Study.} We conducted ablation studies as presented in Table \ref{tab:ablation}. Removing either of the two main stages causes a performance drop, demonstrating Frame-Level Budgeting is essential for eliminating irrelevant frames, while Token-Level Budgeting enhances reasoning by removing redundant tokens. Similarly, ablating key internal mechanisms like proportional budget allocation, Core Token Allocation, or Context Token Allocation also degrades performance. This confirms the necessity of both high-relevance (Core) and diverse (Context) tokens for robust video reasoning.

% \textbf{Sensitivity Analysis.} We further evaluate robustness through sensitivity analysis of key hyperparameters (Figure \ref{fig:hypermeter_analysis}). In each case, we vary one parameter at a time while keeping others fixed at their default values. For the frame scoring weights, we fix the visual dynamic weights to analyze the balance between visual dynamics and query relevance. Notably, across the wide range of tested values for all hyperparameters, the performance consistently remained superior to the corresponding vanilla baseline. Overall, the performance curves for the primary parameters exhibit considerable stability around default values. We also analyzed the number of MMR seed tokens ($k_s$) and found that performance changed minimally for $k_s$ values between 3 and 6. Collectively, these findings indicate that the framework's performance is not highly sensitive to the precise settings of its key hyperparameters, achieving competitive results across a range of values.
% Figure \ref{fig:tradeoff} visualizes the performance-efficiency trade-off of Triage on the LLaVA-OneVision-7B model. While inference time is substantially reduced at lower token retention ratios, performance across the four benchmarks remains highly competitive. This demonstrates the efficacy of our hierarchical budgeting mechanism: by strategically filtering redundant or irrelevant visual information, Triage allows the model to focus its finite computational resources on the most salient information, thereby preserving and even enhancing its reasoning capabilities.

\textbf{Performance-efficiency Trade-off.} Figure \ref{fig:res} visualizes the performance-efficiency trade-off of Triage on the LLaVA-OneVision-7B model. While inference time is substantially reduced at lower token retention ratios, performance across the four benchmarks remains highly competitive. This demonstrates Triage can filter redundant or irrelevant visual information, allowing the model to focus its finite computational resources on the most salient information, thereby preserving and even enhancing its reasoning capabilities.

\textbf{Sensitivity Analysis.} We further evaluate Triage's robustness through sensitivity analysis of key hyperparameters (Figure \ref{fig:res}). In each case, one parameter is varied while others remain fixed. For frame scoring weights, we adjust the balance between visual dynamics and query relevance. We also observed similar robustness for $w_c$ and $w_m$, their plots are omitted due to space limits. Across all tested ranges, the performance consistently exceeded the vanilla baseline, showing that Triage exhibits insensitivity to hyperparameter choices.

\section{Conclusion}
We introduced Triage, a training-free, plug-and-play framework that tackles redundancy in video VLMs. Through hierarchical visual budgeting, Triage selects keyframes and allocates tokens between high-relevance Core Tokens and diverse Context Tokens. Experiments demonstrate that it improves efficiency while maintaining or surpassing baseline performance, providing a scalable solution for video reasoning.

\section{Acknowledgments}
This work was sponsored by the Shenzhen-Hong Kong Joint Funding Project (Category A) under Grant No. SGDX2024\-0115103359001, the National Key Research and Development Program of China under Grant No.2023YFB4502701, Shandong Provincial Natural Science Foundation under Grant No. ZR2024LZH004.

\bibliographystyle{IEEEbib}
\bibliography{strings,refs2}

@inproceedings{radford2021learning,
  title={Learning transferable visual models from natural language supervision},
  author={Radford, Alec and Kim, Jong Wook and Hallacy, Chris and Ramesh, Aditya and Goh, Gabriel and Agarwal, Sandhini and Sastry, Girish and Askell, Amanda and Mishkin, Pamela and Clark, Jack and others},
  booktitle={International conference on machine learning},
  pages={8748--8763},
  year={2021},
  organization={PmLR}
}

@inproceedings{liu2024improved,
  title={Improved baselines with visual instruction tuning},
  author={Liu, Haotian and Li, Chunyuan and Li, Yuheng and Lee, Yong Jae},
  booktitle={Proceedings of the IEEE/CVF conference on computer vision and pattern recognition},
  pages={26296--26306},
  year={2024}
}

@misc{cheng2024videollama2advancingspatialtemporal,
      title={VideoLLaMA 2: Advancing Spatial-Temporal Modeling and Audio Understanding in Video-LLMs}, 
      author={Zesen Cheng and others},
      year={2024},
      journal = {arXiv preprint arXiv:2406.07476},
      url={https://arxiv.org/abs/2406.07476}, 
}

@misc{tao2025dycokedynamiccompressiontokens,
      title={DyCoke: Dynamic Compression of Tokens for Fast Video Large Language Models}, 
      author={Keda Tao and Can Qin and Haoxuan You and Yang Sui and Huan Wang},
      year={2025},
      eprint={2411.15024},
      archivePrefix={arXiv},
      primaryClass={cs.CV},
      url={https://arxiv.org/abs/2411.15024}, 
}

@misc{wang2025adaretakeadaptiveredundancyreduction,
      title={AdaReTaKe: Adaptive Redundancy Reduction to Perceive Longer for Video-language Understanding}, 
      author={Xiao Wang and Qingyi Si and Jianlong Wu and Shiyu Zhu and Li Cao and Liqiang Nie},
      year={2025},
      eprint={2503.12559},
      archivePrefix={arXiv},
      primaryClass={cs.CV},
      url={https://arxiv.org/abs/2503.12559}, 
}

@misc{liu2025boltboostlargevisionlanguage,
      title={BOLT: Boost Large Vision-Language Model Without Training for Long-form Video Understanding}, 
      author={Shuming Liu and Chen Zhao and Tianqi Xu and Bernard Ghanem},
      year={2025},
      eprint={2503.21483},
      archivePrefix={arXiv},
      primaryClass={cs.CV},
      url={https://arxiv.org/abs/2503.21483}, 
}

@misc{xing2025pyramiddropacceleratinglargevisionlanguage,
      title={PyramidDrop: Accelerating Your Large Vision-Language Models via Pyramid Visual Redundancy Reduction}, 
      author={Long Xing and Qidong Huang and Xiaoyi Dong and Jiajie Lu and Pan Zhang and Yuhang Zang and Yuhang Cao and Conghui He and Jiaqi Wang and Feng Wu and Dahua Lin},
      year={2025},
      eprint={2410.17247},
      archivePrefix={arXiv},
      primaryClass={cs.CV},
      url={https://arxiv.org/abs/2410.17247}, 
}

@inproceedings{carbonell1998use,
  title={The use of MMR, diversity-based reranking for reordering documents and producing summaries},
  author={Carbonell, Jaime and Goldstein, Jade},
  booktitle={Proceedings of the 21st annual international ACM SIGIR conference on Research and development in information retrieval},
  pages={335--336},
  year={1998}
}

@misc{wang2024qwen2vlenhancingvisionlanguagemodels,
      title={Qwen2-VL: Enhancing Vision-Language Model's Perception of the World at Any Resolution}, 
      author={Peng Wang and Shuai Bai and Sinan Tan and Shijie Wang and Zhihao Fan and Jinze Bai and Keqin Chen and Xuejing Liu and Jialin Wang and Wenbin Ge and Yang Fan and Kai Dang and Mengfei Du and Xuancheng Ren and Rui Men and Dayiheng Liu and Chang Zhou and Jingren Zhou and Junyang Lin},
      year={2024},
      eprint={2409.12191},
      archivePrefix={arXiv},
      primaryClass={cs.CV},
      url={https://arxiv.org/abs/2409.12191}, 
}

@misc{li2024llavaonevisioneasyvisualtask,
      title={LLaVA-OneVision: Easy Visual Task Transfer}, 
      author={Bo Li and Yuanhan Zhang and Dong Guo and Renrui Zhang and Feng Li and Hao Zhang and Kaichen Zhang and Peiyuan Zhang and Yanwei Li and Ziwei Liu and Chunyuan Li},
      year={2024},
      eprint={2408.03326},
      archivePrefix={arXiv},
      primaryClass={cs.CV},
      url={https://arxiv.org/abs/2408.03326}, 
}

@misc{zhang2024llavavideoinstructiontuningsynthetic,
      title={Video Instruction Tuning With Synthetic Data}, 
      author={Yuanhan Zhang and Jinming Wu and Wei Li and Bo Li and Zejun Ma and Ziwei Liu and Chunyuan Li},
      year={2024},
      eprint={2410.02713},
      archivePrefix={arXiv},
      primaryClass={cs.CV},
      url={https://arxiv.org/abs/2410.02713}, 
}

@article{wu2024longvideobench,
  title={Longvideobench: A benchmark for long-context interleaved video-language understanding},
  author={Wu, Haoning and Li, Dongxu and Chen, Bei and Li, Junnan},
  journal={Advances in Neural Information Processing Systems},
  volume={37},
  pages={28828--28857},
  year={2024}
}

@misc{wang2024lvbenchextremelongvideo,
      title={LVBench: An Extreme Long Video Understanding Benchmark}, 
      author={Weihan Wang and Zehai He and Wenyi Hong and Yean Cheng and Xiaohan Zhang and Ji Qi and Xiaotao Gu and Shiyu Huang and Bin Xu and Yuxiao Dong and Ming Ding and Jie Tang},
      year={2024},
      eprint={2406.08035},
      archivePrefix={arXiv},
      primaryClass={cs.CV},
      url={https://arxiv.org/abs/2406.08035}, 
}

@inproceedings{fu2025videomme,
  title={Video-mme: The first-ever comprehensive evaluation benchmark of multi-modal llms in video analysis},
  author={Fu, Chaoyou and Dai, Yuhan and Luo, Yongdong and Li, Lei and Ren, Shuhuai and Zhang, Renrui and Wang, Zihan and Zhou, Chenyu and Shen, Yunhang and Zhang, Mengdan and others},
  booktitle={Proceedings of the Computer Vision and Pattern Recognition Conference},
  pages={24108--24118},
  year={2025}
}

@inproceedings{li2024mvbench,
  title={Mvbench: A comprehensive multi-modal video understanding benchmark},
  author={Li, Kunchang and Wang, Yali and He, Yinan and Li, Yizhuo and Wang, Yi and Liu, Yi and Wang, Zun and Xu, Jilan and Chen, Guo and Luo, Ping and others},
  booktitle={Proceedings of the IEEE/CVF Conference on Computer Vision and Pattern Recognition},
  pages={22195--22206},
  year={2024}
}

@inproceedings{chen2024fastv,
  title={An image is worth 1/2 tokens after layer 2: Plug-and-play inference acceleration for large vision-language models},
  author={Chen, Liang and Zhao, Haozhe and Liu, Tianyu and Bai, Shuai and Lin, Junyang and Zhou, Chang and Chang, Baobao},
  booktitle={European Conference on Computer Vision},
  pages={19--35},
  year={2024},
  organization={Springer}
}

@misc{liu2025videocompressioncommanderplugandplay,
      title={Video Compression Commander: Plug-and-Play Inference Acceleration for Video Large Language Models}, 
      author={Xuyang Liu and Yiyu Wang and Junpeng Ma and Linfeng Zhang},
      year={2025},
      eprint={2505.14454},
      archivePrefix={arXiv},
      primaryClass={cs.CV},
      url={https://arxiv.org/abs/2505.14454}, 
}

@inproceedings{ding2025dolanguage, series={WWW ’25},
   title={Do Language Models Understand Time?},
   url={http://dx.doi.org/10.1145/3701716.3717744},
   DOI={10.1145/3701716.3717744},
   booktitle={Companion Proceedings of the ACM on Web Conference 2025},
   publisher={ACM},
   author={Ding, Xi and Wang, Lei},
   year={2025},
   month=may, pages={1855–1868},
   collection={WWW ’25} }

@misc{shinde2025surveyefficientvisionlanguagemodels,
      title={A Survey on Efficient Vision-Language Models}, 
      author={Gaurav Shinde and Anuradha Ravi and Emon Dey and Shadman Sakib and Milind Rampure and Nirmalya Roy},
      year={2025},
      eprint={2504.09724},
      archivePrefix={arXiv},
      primaryClass={cs.CV},
      url={https://arxiv.org/abs/2504.09724}, 
}

@article{li2025videochat,
  author    = {K. Li and others},
  title     = {{VideoChat}: Chat-Centric Video Understanding},
  journal   = {Science China Information Sciences},
  volume    = {68},
  number    = {10},
  pages     = {200102},
  year      = {2025},
}

@inproceedings{tu2025vlcache,
  title     = {{VL-Cache}: Sparsity and Modality-Aware {KV} Cache Compression for Vision-Language Model Inference Acceleration},
  author    = {D. Tu and D. Vashchilenko and Y. Lu and P. Xu},
  booktitle = {Proceedings of the International Conference on Learning Representations},
  year      = {2025}
}

\end{document}